\documentclass[10pt,twocolumn,letterpaper]{article}

\usepackage{iccv}
\usepackage{times}
\usepackage{epsfig}
\usepackage{graphicx}
\usepackage{amsmath}
\usepackage{amssymb}
\usepackage{multirow}

\usepackage{array}
\makeatletter
\newcommand{\thickhline}{%
    \noalign {\ifnum 0=`}\fi \hrule height 1pt
    \futurelet \reserved@a \@xhline 
}
\newcolumntype{!}{@{\hskip\tabcolsep\vrule width 1pt\tabcolsep\hskip}} 
\makeatother

\usepackage[breaklinks=true,bookmarks=false]{hyperref}

\iccvfinalcopy 

\def\iccvPaperID{****} 

\DeclareMathOperator*{\argmax}{argmax}

\begin{document}

\title{
\ Bin-wise Temperature Scaling (BTS): Improvement in Confidence Calibration Performance through Simple Scaling Techniques
}

\author{
Byeongmoon Ji \thanks{Equal contribution, alphabetical order} \qquad
Hyemin Jung \footnotemark[1] \qquad
Jihyeun Yoon \qquad
Kyungyul Kim \qquad
Younghak Shin \thanks{Corresponding author} \\
LG CNS\\
Seoul, Korea\\
{\tt\small \{jibm, sweetdream, jihyeun.yoon, kyungyul.kim, younghak.shin\}@lgcns.com}
}

\maketitle

\begin{abstract}
   The prediction reliability of neural networks is important in many applications. Specifically, in safety-critical domains, such as cancer prediction or autonomous driving, a reliable confidence of model’s prediction is critical for the interpretation of the results. Modern deep neural networks have achieved a significant improvement in performance for many different image classification tasks. However, these networks tend to be poorly calibrated in terms of output confidence. Temperature scaling is an efficient post-processing-based calibration scheme and obtains well calibrated results. In this study, we leverage the concept of temperature scaling to build a sophisticated  bin-wise scaling. Furthermore, we adopt augmentation of validation samples for elaborated scaling. The proposed methods consistently improve calibration performance with various datasets and deep convolutional neural network models.
\end{abstract}

\begin{figure*}[ht!]
\begin{center}
   \includegraphics[width=1.0\linewidth]{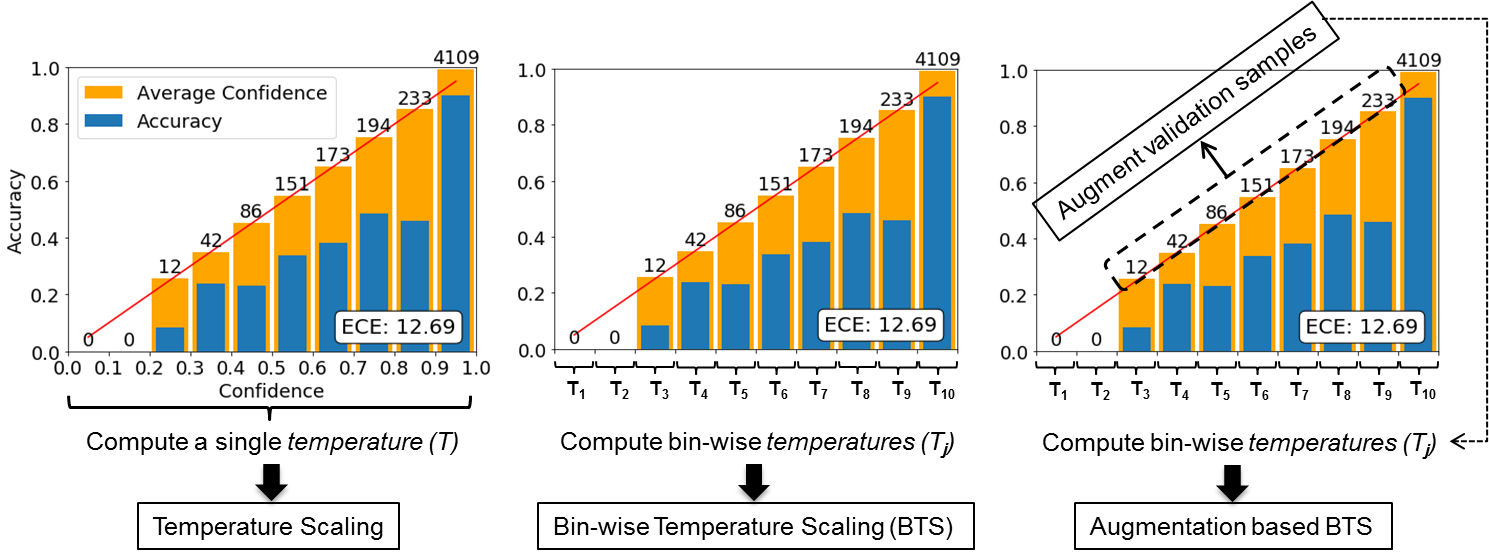}
\end{center}
   \caption{Concepts of TS, BTS, and ABTS using reliability diagrams with the number of samples for each bin. (Left) TS finds a scalar temperature using a validation set and applies it to all test sets. (Center) In BTS, bin-wise temperatures are computed, and each temperature is applied to the corresponding bin-wise test samples. (Right) In ABTS, the validation set is augmented, and the same BTS procedure is conducted.}
\label{figure1}
\end{figure*}

\section{Introduction}
The recent progress made in deep convolutional neural networks (CNNs) has significantly improved their performance in various computer vision tasks, including image classification. For a typical ImageNet classification task~\cite{ImageNet}, recently developed deep CNNs are already performing better than humans~\cite{ResNet}~\cite{DenseNet}. However, as shown in ~\cite{Calibration}, there is a problem in that as the layers of recent networks become deeper, those networks' confidence values (i.e., probabilities for predicted class) become increasingly over-confident. That is, although the accuracy of the latest deep models has been improved, their confidence is much higher than their accuracy.

This can pose significant challenges to interpretation of model's output. For example, in medical applications such as cancer prediction, there is a tremendous difference in the explanation of two diagnoses (each with an incorrect prediction) when one model outputs an overestimated 90\% confidence and the other model outputs a 55\% confidence. Therefore, calibrating a model's output confidence is a critical issue in terms of model reliability and interpretation.

Various confidence calibration methods have recently been proposed in the field of deep learning to overcome the over-confidence issue. Calibration studies can be divided into two main types. The first is a post-processing-based calibration method using an already trained model~\cite{Calibration}, and the second method performs calibration while training the model~\cite{Regularizing}~\cite{Stochastic}~\cite{KernelMean}.

The most simple and effective calibration method is a post-processing technique called \textit{temperature scaling} (TS)~\cite{Calibration}. The authors showed that the predicted confidence of recent CNNs is not calibrated, and they proposed simple post-processing-based scaling using the validation dataset. This method calibrates predicted confidence well for various image datasets. The advantage of this method is its ability to calibrate confidence without affecting the test error rate.

In ~\cite{Regularizing}, the authors proposed several methods to control confidence during model training. They showed that directly penalizing confidence using an entropy measure during training is an effective technique in terms of generalization performance. However, calibration performance was not evaluated in their study. A variance-weighted confidence-integrated loss function was proposed for calibrating confidence during model training in ~\cite{Stochastic}. Similarly, in ~\cite{KernelMean}, the authors proposed a method for conducting calibration during training. They included a direct term for minimizing the calibration error into the training loss. Both methods ~\cite{Stochastic}~\cite{KernelMean} have the advantage that a separate validation dataset is not needed. However, compared to temperature scaling, calibration performance in ~\cite{Stochastic} and ~\cite{KernelMean} was not consistently improved.

Because simple post-processing-based calibration strategies are powerful tools for calibrating confidence without diminishing the test error rate, we focus on post-processing-based calibration techniques in this study. We propose the use of bin-wise temperature scaling (BTS), which aims to find multiple temperatures for different confidence ranges. In addition, we suggest augmentation-based bin-wise temperature scaling (ABTS) for finding stable bin temperatures. We evaluate the proposed methods using several recent deep models, including ResNet~\cite{ResNet} and DenseNet~\cite{DenseNet}, and obtain outstanding calibration performance on various image classification datasets, such as CIFAR-10/100~\cite{CIFAR}, Caltech Birds~\cite{Birds}, and Stanford Cars~\cite{Cars}.


\section{Methods}
Consider the multi-class ($C > 2$) classification problem with inputs $X = \{\mathbf{x_1}, \mathbf{x_2}, \ldots, \mathbf{x_i}\}$ and corresponding class labels $Y = \{ y_1, y_2, \ldots, y_i \}$. For an instance $\mathbf{x_i}$ (with corresponding label $y_i$), the model $\phi$ predicts the label $\hat{y_i} = \argmax \phi(y|\mathbf{x_i})$. The confidence score $\hat{p_i}$ is defined as 

\begin{equation}
  \hat{p_i} = \max\limits_c \sigma_{SM}(\mathbf{z_i})^c,
\end{equation}
where $\mathbf{z_i}$ is the logits vector for input $\mathbf{x_i}$, and $\sigma_{SM}$ represents the softmax function. A well-calibrated confidence $\hat{p}$ should reflect a true probability. For example, if we collect ten samples, each having an identical confidence score of 0.8, we then expect an 80\% classification accuracy for the ten samples.

In Figure 1, we visualize the baseline TS concept and the two proposed approaches, BTS and ABTS, using a reliability diagram~\cite{Calibration}~\cite{RD1}. We describe in detail the procedure for each technique in the following subsections.


\subsection{Temperature Scaling}
With TS~\cite{Calibration}, a scalar temperature $t$ is determined using a validation set, and it is applied to all test samples for confidence calibration. Formally, the calibrated confidence score $\hat{q}$ is defined as 
\begin{equation}
  \hat{q_i} = \max\limits_c \sigma_{SM}(\mathbf{z_i}/t)^c.
\end{equation}
Here, $t$ is a positive scalar parameter (temperature) that aims to soften softmax in the case of overconfidence with $t > 1$. The temperature $t$ is optimized by minimizing the negative log likelihood (NLL) loss on the validation set~\cite{Calibration}.

With TS, a single $t$ is computed and applied to all test samples. However, normally most samples have a high confidence as shown in Figure 1; therefore, temperature $t$ could easily be biased toward high-confidence samples, and might not be optimal for low-confidence samples. Motivated by this, we propose the use of bin-wise temperature scaling methods.


\subsection{Bin-wise Temperature Scaling}

In BTS, we divide the samples into multiple bins. Then, using the validation samples of each bin, bin-wise temperatures are computed. In each test sample, the corresponding temperature is applied for scaling based on the test confidence as follows:
\begin{equation}
  \hat{q_i} = \max\limits_c \sigma_{SM}(\mathbf{z_i^j}/t^j)^c,
\end{equation}
where $j$ is the bin number corresponding to the test confidence.

Confidence-interval-based binning is a simple and intuitive binning method. In this case, the confidence range [0, 1] is partitioned into $N$ equal-sized bins. This simple binning method has a problem that temperatures with a low confidence are computed using an extremely small number of validation samples, as shown in Figure 1. Therefore, this method tends to be unstable for the corresponding test samples.

To overcome this issue, we propose a binning method based on the number of samples. With this method, we collect the same number of validation samples in each bin— except for high-confidence samples. If we had collected the same number of samples from the crowded high-confidence area, then extremely high-confidence samples (confidence greater than 0.999) with small differences (under 0.0001) would have been placed into many different bins, resulting in redundant temperatures across test samples with almost the same confidence. Therefore, we utilize a 0.999 threshold for collecting high-confidence samples. Thus, if validation samples have a confidence greater than 0.999, we collect them in a single bin, regardless of the number of samples.


\subsection{Augmentation based Bin-wise Temperature Scaling}
As mentioned in Section 2.2, temperatures for low-confidence parts may be unstable when applying a confidence-interval-based binning method. The proposed binning method (based on the number of samples) has a similar weak point. Because a small number of validation samples are positioned in low-confidence areas, the range of confidence of such areas increases if we collect the same number of samples in each bin. 

To overcome this issue, we adopt an image augmentation technique. Image augmentation (e.g., image rotation, image shifting, and brightness adjustment) is widely used in the training phase of deep neural networks. It is known that such augmentation is efficient when the number of training samples is limited. In recent studies~\cite{TestAug1}~\cite{TestAug2}, simple augmentation techniques have been adopted to test samples and improve model performance during the testing period. 

We expect that augmenting the validation samples will help identify stable temperatures for low-confidence bins and further improve the performance of BTS. In the ABTS method, as shown in Figure 1 (right), we first apply an image augmentation technique to bins that have a small number of validation samples. The optimal number of bins to augment can differ by dataset and model. Instead of optimizing this number, we choose eight bins with confidence less than 0.8 for all datasets and models in this study. 

After augmentation, we perform the same procedure with BTS. We evaluate widely used image augmentation techniques (such as image shifting, brightness adjustment, contrast control, and adding blur) with ABTS in our experiment. We apply each augmentation technique to the validation samples of eight bins; therefore, the number of samples is doubled after each augmentation.


\section{Experiments}

In this study, we compare the calibration performance of the proposed BTS and ABTS methods with conventional TS using a prevalent calibration measure, expected calibration error (ECE)~\cite{Calibration}~\cite{ECE}. In addition, we evaluate four image datasets using five network architectures.


\subsection{Datasets}

\begin{enumerate}
    \item CIFAR-10/100~\cite{CIFAR}: Color images ($32 \times 32$ pixel resolution) from 10/100 classes were applied: 45,000, 5,000, and 10,000 images for the training, validation, and testing sets, respectively. The original dataset included 50,000 and 10,000 training and testing images, respectively. Therefore, we randomly divide the training set into training and validation sets. 
    \item Caltech-UCSD Birds~\cite{Birds}: Color images of 200 bird species drawn from ImageNet were applied: 5,994, 2,897, and 2,897 images for the training, validation, and testing sets, respectively.
    \item Stanford Cars~\cite{Cars}: Color images of 196 classes of car by make, model, and year were applied: 8,041, 4,020, and 4,020 images for the training, validation, and testing sets, respectively.
\end{enumerate}


\subsection{Models}
For CIFAR-10/100, we use four well-known CNNs: ResNet-50/110~\cite{ResNet}, Wide ResNet 28-10~\cite{WideResNet}, DenseNet 100~\cite{DenseNet}, and VGG 16~\cite{VGGNet}. For the Caltech-UCSD Birds and Stanford Cars datasets, we use two CNNs: DenseNet and ResNet-101~\cite{ResNet}. To train the classification models, we normalize all images using the pixel mean and standard deviation, and we apply the stochastic gradient decent method with a momentum of 0.9 for training. We utilize other hyperparameters, such as the initial learning rate and learning rate schedule, as described in the above studies, for training each model. We train CIFAR-10 and CIFAR-100 from scratch and use the ImageNet pretrained weights for the Caltech-UCSD Birds and Stanford Cars datasets.


\begin{table*}[h!]
\centering
\resizebox{\textwidth}{!}{%
\begin{tabular}{|c|c|c|c|c|c|c|c|c|}
\hline
\multirow{4}{*}{Dataset} & \multirow{4}{*}{Model} & \multirow{4}{*}{\begin{tabular}[c]{@{}c@{}}Test error\\ (\%)\end{tabular}} & \multicolumn{6}{c|}{ECE (\%)} \\ \cline{4-9} 
 &  &  & \multirow{3}{*}{\begin{tabular}[c]{@{}c@{}}Baseline\\ (uncalibrated)\end{tabular}} & \multirow{3}{*}{TS} & \multicolumn{2}{c|}{BTS} & \multicolumn{2}{c|}{ABTS} \\ \cline{6-9} 
 &  &  &  &  & \begin{tabular}[c]{@{}c@{}}Confidence\\ Interval\end{tabular} & \begin{tabular}[c]{@{}c@{}}Number of\\ Samples\end{tabular} & \begin{tabular}[c]{@{}c@{}}Confidence\\ Interval\end{tabular} & \begin{tabular}[c]{@{}c@{}}Number of\\ Samples\end{tabular} \\ \hline \hline
\rule{0pt}{2ex}CIFAR-10 & VGG 16 & 7.10 & 4.99 & 1.45 & 1.26 & 0.90 & \textit{0.73} & \textbf{0.37} \\
\rule{0pt}{2ex}CIFAR-10 & ResNet 110 & 6.23 & 4.37 & 0.98 & 1.12 & 0.79 & \textit{0.58} & \textbf{0.36} \\
\rule{0pt}{2ex}CIFAR-10 & DenseNet 100(k=12) & 5.37 & 3.09 & 1.05 & 0.93 & 0.87 & \textbf{0.53} & \textit{0.69} \\
\rule{0pt}{2ex}CIFAR-10 & Wide ResNet 28-10 & 4.80 & 3.03 & 0.73 & 0.78 & \textit{0.55} & 0.64 & \textbf{0.44} \\
\rule{0pt}{2ex}CIFAR-100 & VGG 16 & 26.56 & 14.13 & 5.20 & 4.92 & 4.98 & \textit{3.41} & \textbf{2.53} \\
\rule{0pt}{2ex}CIFAR-100 & ResNet 110 & 28.51 & 15.79 & 2.54 & 2.32 & 2.28 & \textbf{0.99} & \textit{1.29} \\
\rule{0pt}{2ex}CIFAR-100 & DenseNet 100 (k=12) & 25.11 & 10.29 & 3.55 & 2.53 & 2.37 & \textbf{1.55} & \textit{1.64} \\
\rule{0pt}{2ex}CIFAR-100 & Wide ResNet 28-10 & 21.05 & 7.16 & 5.01 & 3.08 & 2.97 & \textit{1.96} & \textbf{1.74} \\
\rule{0pt}{2ex}Birds & ResNet 101 & 21.44 & 2.13 & 2.27 & 1.33 & 1.40 & \textit{1.28} & \textbf{1.21} \\
\rule{0pt}{2ex}Birds & DenseNet 100 (k=12) & 19.92 & 3.22 & 1.44 & \textit{1.42} & 1.57 & 1.70 & \textbf{1.40} \\
\rule{0pt}{2ex}Cars & ResNet 101 & 12.16 & 2.06 & 2.05 & 1.24 & 1.36 & \textbf{0.96} & \textit{1.16} \\
\rule{0pt}{2ex}Cars & DenseNet 100 (k=12) & 9.43 & 2.31 & 2.20 & 1.24 & 1.16 & \textbf{0.89} & \textit{1.09} \\ \hline
\end{tabular}%
}
\caption{ECE (\%) results of the baseline model, TS, and the proposed methods (BTS and ABTS) on different datasets and models. The best and second-best ECE values are represented in bold and italics, respectively. For augmentation with ABTS, we applied image shifting along the x-axis within a range of [-4, -8] on CIFAR-10 and 100 and [-28, -56] on Caltech-UCSD Birds and Stanford Cars dataset based on the size of the original images.}
\end{table*}

\subsection{Evaluation Metrics}
As shown in Figures 1 and 2, we utilize a reliability diagram~\cite{Calibration}~\cite{RD1}~\cite{RD2} for visualizing calibration performance. A reliability diagram is the most intuitive and readable way to visualize calibration performance. In a reliability diagram, test samples are binned based on the range of confidence; i.e., samples with a similar confidence are collected into the same bin. The reliability diagram then shows the accuracy and average confidence of each bin. If the model is perfectly calibrated, the accuracy and average confidence should be the same.

To numerically evaluate calibration performance, ECE is the most widely used evaluation metric~\cite{Calibration}~\cite{VGGNet}. Here, $B = \{B_1, B_2, \ldots, B_N\}$ represents $N$ different bins based on confidence range, and $ACC(B_j)$ and $CONF(B_j)$ are the accuracy and average confidence of each bin $B_j$, respectively. ECE is the weighted average of the difference between accuracy and average confidence for each bin.
\begin{equation}
  ECE = {1 \over n} \sum\limits^N_{j=1} \|B_j \| \cdot \|ACC(B_j) - CONF(B_j) \|.
\end{equation}


\begin{figure*}[h!]
\begin{center}
   \includegraphics[width=1.0\linewidth]{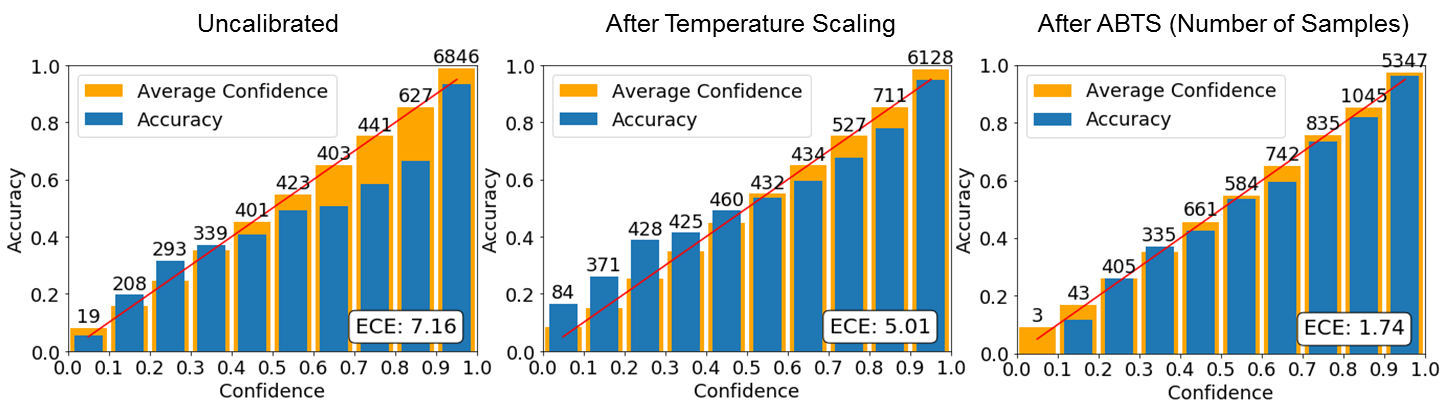}
\end{center}
   \caption{Reliability diagrams and ECE (\%) of (left) baseline uncalibrated model, (center) TS, and (right) proposed ABTS (based on the number of samples binned) for Wide ResNet on CIFAR-100.}
   
\label{figure2}
\end{figure*}

\section{Results}

As shown in Table 1, we evaluate the calibration performance of BTS and ABTS by comparing the test ECE (\%) with an uncalibrated model (as a baseline) and TS across various image datasets and models. With BTS and ABTS, the results of two different binning methods (confidence interval and number of samples) are represented. We use a fixed number of bins (50) for both BTS and ABTS, as shown in Table 1. This number can be changed to obtain better calibration results for each dataset and model (see Table 3).

As shown in Table 1, the simple BTS method achieves better calibration results compared to TS for most of the models and datasets; that is, for the binning methods based on the confidence interval and number of samples, BTS shows improved calibration performance compared to TS in 10 of 12 and 11 of 12 cases, respectively. In addition, the best and second-best (in bold and italic, respectively) calibration performances are generally obtained using ABTS based on the number of samples and the confidence interval, respectively. Specifically, ABTS based on the number of samples (last column) outperforms TS for all experiment cases listed in Table 1. Note that TS, BTS, and ABTS are all post-processing-based scaling techniques; therefore, their test error rates are the same as that of the baseline model.

In Figure 2, we aim to analyze the effect of the proposed method by comparing the reliability diagrams of the uncalibrated baseline model (left), TS (center), and the proposed ABTS binning method based on the number of samples (right). For this comparison, we use the results of Wide ResNet on the CIFAR-100 dataset, as shown in Table 1. For ABTS (right), image shifting was applied for augmenting the validation samples. 

With the uncalibrated baseline model (left), samples with a lower confidence (under 0.4) are under-confident, and samples with a higher confidence (over 0.4) are over-confident. In this case, for ideal calibration, a lower confidence should be increased, and a higher confidence should be decreased. However, when applying temperature scaling using a single temperature for all the samples, it is difficult to adopt two different scaling strategies. After temperature scaling (center), a large number of over-confident samples led to a decreased overall confidence and a slightly improved ECE, but low-confidence samples are more under-confident than the baseline. By contrast, because we apply bin-wise temperatures that adjust the scaling used for each bin, after applying ABTS (right), the overall confidence is well-calibrated; therefore, a much lower ECE is obtained compared to TS.

\begin{figure}[h!]
\begin{center}
   \includegraphics[width=1.0\columnwidth]{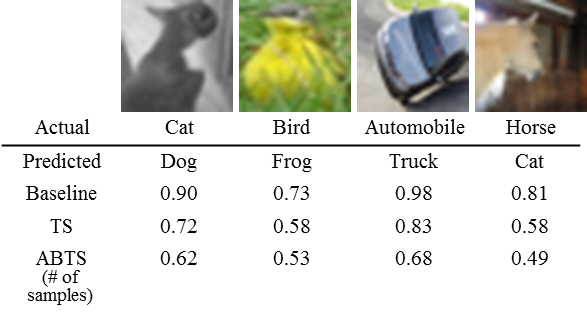}
\end{center}
   \caption{Example pictures from CIFAR-10 test set with actual class labels and those predicted by the VGG 16 model. Confidence values are listed for baseline, TS, and ABTS using the number of samples. For ABTS, image shifting was applied for augmenting the validation samples.}
   
\label{figure3}
\end{figure}

\begin{table}[h!]
\centering
\resizebox{\columnwidth}{!}{%
\begin{tabular}{|c|c|c|c|c|c|}
\hline
Model & \begin{tabular}[c]{@{}c@{}}Augmen-\\ tation\end{tabular} & Baseline & TS & \begin{tabular}[c]{@{}c@{}}ABTS \\   (CI)\end{tabular} & \begin{tabular}[c]{@{}c@{}}ABTS \\   (\# of\\ samples)\end{tabular} \\ \hline \hline
\multirow{4}{*}{VGG 16} & Shift & \multirow{4}{*}{14.13} & \multirow{4}{*}{5.20} & 3.41 & 2.53 \\
 & Bright &  &  & 3.35 & 2.76 \\
 & Contrast &  &  & 3.54 & 2.51 \\
 & Blur &  &  & 3.80 & 3.10 \\ \hline
\multirow{4}{*}{ResNet 110} & Shift & \multirow{4}{*}{15.79} & \multirow{4}{*}{2.54} & 0.99 & 1.20 \\
 & Bright &  &  & 1.36 & 1.40 \\
 & Contrast &  &  & 1.07 & 1.68 \\
 & Blur &  &  & 1.23 & 1.76 \\ \hline
\multirow{4}{*}{\begin{tabular}[c]{@{}c@{}}DenseNet\\ 100 (k=12)\end{tabular}} & Shift & \multirow{4}{*}{10.29} & \multirow{4}{*}{3.55} & 1.55 & 1.64 \\
 & Bright &  &  & 1.16 & 1.14 \\
 & Contrast &  &  & 1.54 & 1.47 \\
 & Blur &  &  & 1.69 & 1.51 \\ \hline
\multirow{4}{*}{\begin{tabular}[c]{@{}c@{}}Wide ResNet\\ 28-10\end{tabular}} & Shift & \multirow{4}{*}{7.16} & \multirow{4}{*}{5.01} & 1.96 & 1.74 \\
 & Bright &  &  & 1.81 & 1.78 \\
 & Contrast &  &  & 1.58 & 1.35 \\
 & Blur &  &  & 1.36 & 1.10 \\ \hline
\end{tabular}%
}
\caption{ECE (\%) comparison of different augmentations in ABTS methods based on the confidence interval (CI) and number of samples on CIFAR-100. }
\end{table}

Figure 3 shows four example pictures of CIFAR-10 test set, each with its actual and predicted class labels. The predicted confidence (i.e., maximum confidence) values from the baseline VGG 16 model along with those obtained using TS and ABTS based on the number of samples are listed. These four pictures are incorrectly classified by the model and also appear ambiguous to human eyes. However, the confidence values of the baseline model are unnecessarily high for the incorrect labels. This may result in an unreliable interpretation of the its output. For example, the third image's actual class label is “Automobile,” but the baseline model incorrectly classified it as “Truck” with a very high confidence (0.98). For the same image, after confidence calibration, TS and ABTS exhibit a reduced confidence of 0.83 and 0.68, respectively. The proposed ABTS method not only shows the lowest ECE as listed in Table 1 (CIFAR-10 and VGG 16) but also exhibits the lowest confidence (i.e. most uncertainty) when incorrectly predicting the ambiguous images as shown in Figure 3.

In the results of Table 1, we use one type of augmentation, image shifting, for the proposed ABTS methods. Typically, a variety of image augmentation techniques are adopted to augment the training samples in image classification tasks. We aim to evaluate the calibration performance of the proposed ABTS methods for different augmentation techniques, as shown in Table 2. We adopt image shifting, brightness adjustment, contrast control, and blur for augmenting the validation samples, and we evaluate these on the CIFAR-100 dataset.

For our image-shifting augmentation, we shift the input image along the x-axis using a random value ranging from -4 to -8. For brightness control, a value ranging from -150 to 150 is randomly selected and added to all pixels. We use linear contrast with a strength parameter of 0.5 for contrast control. For blur augmentation, we apply Gaussian blur with its sigma parameter randomly set between 0 and 1. To implement these augmentations, we use an open-source Python library~\cite{imgaug}. 

As the results in Table 2 indicate, the proposed ABTS methods based on the confidence interval and number of samples show better calibration performance than TS for the four different augmentation techniques. There is no clear leader among the different augmentation techniques, and the difference in performance for each augmentation is marginal in each model. We note that strong data augmentation can change the distribution between the validation and test sets, thereby diminishing the effectiveness of ABTS.

\begin{table}[h!]
\centering
\resizebox{\columnwidth}{!}{%
\begin{tabular}{|c|c|c|c|c|c|}
\hline
Model & Baseline & TS & \begin{tabular}[c]{@{}c@{}}\# of\\ bins\end{tabular} & \begin{tabular}[c]{@{}c@{}}BTS \\   (CI)\end{tabular} & \begin{tabular}[c]{@{}c@{}}ABTS \\   (\# of\\ samples)\end{tabular} \\ \hline \hline
\multirow{4}{*}{VGG 16} & \multirow{4}{*}{14.13} & \multirow{4}{*}{5.20} & 5 & 4.92 & 2.76 \\
 &  &  & 10 & 4.90 & 2.52 \\
 &  &  & 20 & 4.90 & 2.51 \\
 &  &  & 50 & 4.98 & 2.53 \\ \hline
\multirow{4}{*}{ResNet 110} & \multirow{4}{*}{15.79} & \multirow{4}{*}{2.54} & 5 & 2.37 & 1.13 \\
 &  &  & 10 & 2.26 & 1.10 \\
 &  &  & 20 & 2.24 & 1.15 \\
 &  &  & 50 & 2.28 & 1.20 \\ \hline
\multirow{4}{*}{\begin{tabular}[c]{@{}c@{}}DenseNet\\ 100 (k=12)\end{tabular}} & \multirow{4}{*}{10.29} & \multirow{4}{*}{3.55} & 5 & 2.28 & 1.36 \\
 &  &  & 10 & 2.32 & 1.25 \\
 &  &  & 20 & 2.15 & 1.50 \\
 &  &  & 50 & 2.37 & 1.64 \\ \hline
\multirow{4}{*}{\begin{tabular}[c]{@{}c@{}}Wide ResNet\\ 28-10\end{tabular}} & \multirow{4}{*}{7.16} & \multirow{4}{*}{5.01} & 5 & 3.42 & 2.06 \\
 &  &  & 10 & 2.69 & 1.74 \\
 &  &  & 20 & 2.96 & 1.74 \\
 &  &  & 50 & 2.97 & 1.74 \\ \hline
\end{tabular}%
}
\caption{ECE (\%) comparison of different numbers of bins in BTS and ABTS on CIFAR-100. Image-shifting augmentation is used for ABTS.}
\end{table}

For the proposed BTS and ABTS methods shown in Tables 1 and 2, we fix the number of bins at fifty, although the optimal number of bins might differ for different datasets and models. In Table 3, we compare the ECE for different numbers of bins (e.g., 5, 10, 20, and 50) for BTS and ABTS (based on the number of samples) on CIFAR-100. The exact number of bins may be smaller owing to the use of a threshold for high-confidence samples in number-of-samples-based binning (see Section 2.2). The results indicate that, for all models, the ECE of the proposed BTS and ABTS methods is not sensitive to differing numbers of bins. Thus, regardless of the number of bins used, the proposed methods outperform the baseline TS method.


\section{Conclusion}

Calibrating the output confidence of a neural network is essential for reliably interpreting its results in many applications. Post-processing-based TS was shown to be the simplest and most effective calibration scheme, and this scheme does not diminish test error performance. However, TS relies on a single temperature to soften softmax for all samples. In this study, we aimed to propose a more sophisticated post-processing based scaling method. We proposed the use of bin-wise temperature scaling (BTS), which adopts a binning method based on either the confidence interval or the number of samples. In addition, to further improve calibration performance, we suggested the use of an augmentation-based BTS method, which augmented the validation samples to find a stable temperature for each bin. We evaluated these ideas using various deep CNNs and image datasets. The experimental results indicate that the proposed, simple idea outperforms the baseline temperature scaling approach.


{\small
\bibliographystyle{ieee}
\bibliography{egbib}

\begin{thebibliography}{10}\itemsep=-1pt

\bibitem{RD1}
M.~H. DeGroot and S.~E. Fienberg.
\newblock The comparison and evaluation of forecasters.
\newblock {\em The Statistician: Journal of the Institute of Statisticians},
  32:12--22, 1983.

\bibitem{TestAug2}
I.~Golan and R.~El{-}Yaniv.
\newblock Deep anomaly detection using geometric transformations.
\newblock {\em CoRR}, abs/1805.10917, 2018.

\bibitem{Calibration}
C.~Guo, G.~Pleiss, Y.~Sun, and K.~Q. Weinberger.
\newblock On calibration of modern neural networks.
\newblock {\em CoRR}, abs/1706.04599, 2017.

\bibitem{ResNet}
K.~He, X.~Zhang, S.~Ren, and J.~Sun.
\newblock Deep residual learning for image recognition.
\newblock {\em CoRR}, abs/1512.03385, 2015.

\bibitem{DenseNet}
G.~Huang, Z.~Liu, and K.~Q. Weinberger.
\newblock Densely connected convolutional networks.
\newblock {\em CoRR}, abs/1608.06993, 2016.

\bibitem{imgaug}
A.~B. Jung.
\newblock {imgaug}.
\newblock \url{https://github.com/aleju/imgaug}, 2018.
\newblock [Online; accessed 30-Oct-2018].

\bibitem{Cars}
J.~Krause, M.~Stark, J.~Deng, and L.~Fei-Fei.
\newblock 3d object representations for fine-grained categorization.
\newblock In {\em 4th International IEEE Workshop on 3D Representation and
  Recognition (3dRR-13)}, Sydney, Australia, 2013.

\bibitem{CIFAR}
A.~Krizhevsky, V.~Nair, and G.~Hinton.
\newblock Cifar-100 (canadian institute for advanced research).

\bibitem{KernelMean}
A.~Kumar, S.~Sarawagi, and U.~Jain.
\newblock Trainable calibration measures for neural networks from kernel mean
  embeddings.
\newblock In J.~Dy and A.~Krause, editors, {\em Proceedings of the 35th
  International Conference on Machine Learning}, volume~80 of {\em Proceedings
  of Machine Learning Research}, pages 2805--2814, Stockholmsmässan, Stockholm
  Sweden, 10--15 Jul 2018. PMLR.

\bibitem{ECE}
M.~P. Naeini, G.~F. Cooper, and M.~Hauskrecht.
\newblock Obtaining well calibrated probabilities using bayesian binning.
\newblock In {\em Proceedings of the Twenty-Ninth AAAI Conference on Artificial
  Intelligence}, AAAI'15, pages 2901--2907. AAAI Press, 2015.

\bibitem{RD2}
A.~Niculescu-Mizil and R.~Caruana.
\newblock Predicting good probabilities with supervised learning.
\newblock In {\em Proceedings of the 22Nd International Conference on Machine
  Learning}, ICML '05, pages 625--632, New York, NY, USA, 2005. ACM.

\bibitem{Regularizing}
G.~Pereyra, G.~Tucker, J.~Chorowski, L.~Kaiser, and G.~E. Hinton.
\newblock Regularizing neural networks by penalizing confident output
  distributions.
\newblock {\em CoRR}, abs/1701.06548, 2017.

\bibitem{ImageNet}
O.~Russakovsky, J.~Deng, H.~Su, J.~Krause, S.~Satheesh, S.~Ma, Z.~Huang,
  A.~Karpathy, A.~Khosla, M.~Bernstein, A.~C. Berg, and L.~Fei-Fei.
\newblock {ImageNet Large Scale Visual Recognition Challenge}.
\newblock {\em International Journal of Computer Vision (IJCV)},
  115(3):211--252, 2015.

\bibitem{Stochastic}
S.~Seo, P.~H. Seo, and B.~Han.
\newblock Confidence calibration in deep neural networks through stochastic
  inferences.
\newblock {\em CoRR}, abs/1809.10877, 2018.

\bibitem{VGGNet}
K.~Simonyan and A.~Zisserman.
\newblock Very deep convolutional networks for large-scale image recognition.
\newblock In {\em International Conference on Learning Representations}, 2015.

\bibitem{TestAug1}
R.~Timofte, R.~Rothe, and L.~V. Gool.
\newblock Seven ways to improve example-based single image super resolution.
\newblock {\em CoRR}, abs/1511.02228, 2015.

\bibitem{Birds}
C.~Wah, S.~Branson, P.~Welinder, P.~Perona, and S.~Belongie.
\newblock {The Caltech-UCSD Birds-200-2011 Dataset}.
\newblock Technical Report CNS-TR-2011-001, California Institute of Technology,
  2011.

\bibitem{WideResNet}
S.~Zagoruyko and N.~Komodakis.
\newblock Wide residual networks.
\newblock {\em CoRR}, abs/1605.07146, 2016.

\end{thebibliography}
}

\end{document}